\documentclass[lettersize,journal,captionsoff]{IEEEtran}
\usepackage{amsmath,amsfonts}
\usepackage{algorithmic}
\usepackage{algorithm}
\usepackage{array}
\usepackage[caption=false,font=normalsize,labelfont=sf,textfont=sf]{subfig}
\usepackage{textcomp}
\usepackage{stfloats}
\usepackage{url}
\usepackage{verbatim}
\usepackage{graphicx}
\usepackage{cite}
\usepackage{booktabs}       
\usepackage{subcaption}
\usepackage{multirow}
\usepackage{xcolor}
\usepackage[colorlinks,linkcolor=red,anchorcolor=blue,citecolor=green,urlcolor=black]{hyperref}

\newcommand{\etal}{et al.~}

\begin{document}

\title{Enhancing Robust Representation in Adversarial Training: Alignment and Exclusion Criteria}

\author{
Nuoyan Zhou, 
Nannan Wang*\thanks{* represents the Corresponding Author.},
Decheng Liu, 
Dawei Zhou, 
Xinbo Gao 
}

\markboth{}%
{Shell \MakeLowercase{\textit{et al.}}: A Sample Article Using IEEEtran.cls for IEEE Journals}

\maketitle

\begin{abstract}
Deep neural networks are vulnerable to adversarial noise. Adversarial Training (AT) has been demonstrated to be the most effective defense strategy to protect neural networks from being fooled. However, we find AT omits to learning robust features, resulting in poor performance of adversarial robustness. To address this issue, we highlight two criteria of robust representation: 
(1) Exclusion: \emph{the feature of examples keeps away from that of other classes}; (2) Alignment: \emph{the feature of natural and corresponding adversarial examples is close to each other}. These motivate us to propose a generic framework of AT to gain robust representation, by the asymmetric negative contrast and reverse attention. Specifically, we design an asymmetric negative contrast based on predicted probabilities, to push away examples of different classes in the feature space. Moreover, we propose to weight feature by parameters of the linear classifier as the reverse attention, to obtain class-aware feature and pull close the feature of the same class. Empirical evaluations on three benchmark datasets show our methods greatly advance the robustness of AT and achieve state-of-the-art performance. \footnote{Our code is available at \url{https://github.com/changzhang777/ANCRA}.}
\end{abstract}


\section{Introduction}

\IEEEPARstart{D}{eep} Neural Networks (DNNs) have achieved great success in academia and industry, but they are easily fooled by carefully crafted Adversarial Examples \textbf{(AEs)} to output incorrect results~\cite{FGSM}, which leads to potential threats and insecurity in the application. Given a naturally trained DNN and a natural example, an adversarial example can be generated by adding small perturbations to the natural example. Adversarial examples can always fool models to make incorrect output. At the same time, it is difficult to distinguish adversarial examples from natural examples by human eyes. In recent years, lots of researches reveal adversarial examples can be crafted in various fields, including image classification~\cite{FGSM,PGD,C&W,AA,doan2022tnt,chen2023query,zhang2020walking}, object detection~\cite{ad_for_ss_and_od,ROBUStOD}, natural language processing~\cite{textattack_NLP,NLPattack}, semantic segmentation~\cite{SSattack,adversarialSS}, etc. The vulnerability of DNNs has aroused common concerns on adversarial robustness.


Many defense methods have been proposed to protect DNNs from adversarial perturbations, such as Adversarial Training (AT)~\cite{PGD,TRADES,mart, SAT, MAN, FAT, AWP, kuang2023semantically, lee2023robust}, image denoising~\cite{HGD,li2023transformer,9866824}, defensive distillation~\cite{Multi-teacherAdversarialDistillation,SmoothandDistillation, liu2022mutual,bai2023guided} and so on. Among them, AT has reached excellent robust performance and is universally recognized as one of the most effective defense methods. Existing work~\cite{Multi-teacherAdversarialDistillation,SmoothandDistillation,FAT,TE,SAT,S2O,MAN} has improved the effectiveness of AT in many aspects, but few studies pay attention to learning robust feature. The overlook may lead to potential threats in the feature space of AT models, which harms robust classification. Although some techniques like Adversarial Contrastive Learning (ACL)~\cite{ROCL,ADVCL,A-InfoNCE, AdversarialSCL} and robust feature selection ~\cite{KWTA,CAS,cifs,audio} are committed to optimizing feature distribution, they don't reach a consensus on criteria for robust feature learning. We pose three questions to investigate the connection between robust representation and AT, and attempt to enhance AT through robust representation learning.

\emph{Q. 1 Does the absence of robust representation learning in AT result in a deficient feature distribution?}

\begin{figure}[t]
    \vspace{-0.0em}
    \centering
    \includegraphics[width=0.5\textwidth]{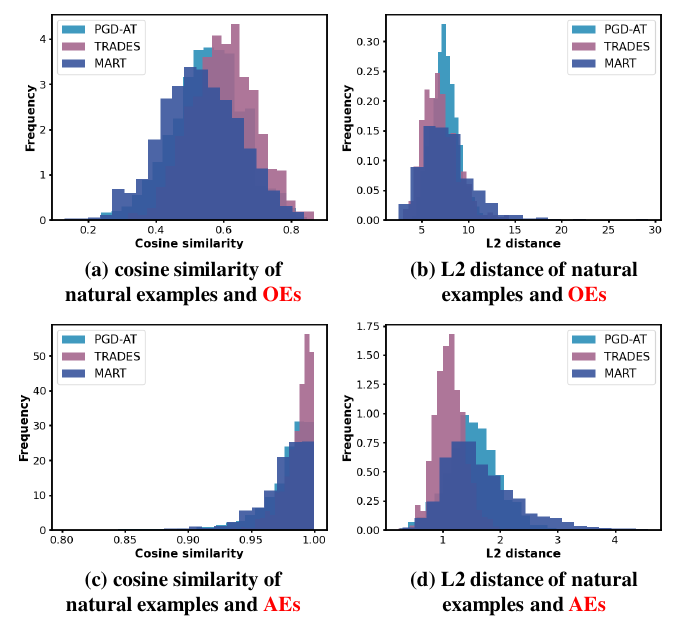}
    \caption{Frequency histograms of the {$L_2$} distance and cosine similarity of the feature of natural examples, AEs and \textbf{O}ther classes' \textbf{E}xamples \textbf{(OEs)}. (a) and (b) show a part of examples with different labels are quite similar to each other (cosine similarity $\geq$ 0.6, {$L_2$} $\leq$ 5.0), (c) and (d) show a few natural examples are far from their adversarial examples though we expect they have the same predicted class (cosine similarity $\leq$ 0.95, {$L_2$} $\geq$ 2.0). These indicate the feature of AT is defective.}
    \label{fig:AT_f}
\end{figure}

To demonstrate AT is indeed deficient in the representation, We train ResNet-18 models on CIFAR-10 with benchmark AT methods: PDG-AT~\cite{PGD}, TRADES~\cite{TRADES}, MART~\cite{mart}. We measure the distance and similarity of the feature between natural examples, AEs and \textbf{O}ther classes' \textbf{E}xamples \textbf{(OEs)}. As shown in Figure~\ref{fig:AT_f} (a) and Figure~\ref{fig:AT_f} (b), the cosine similarity of natural examples and OEs shows a bell-shaped distribution between 0.1 and 0.5, and the $L_2$ distance shows a skewed distribution between 2.0 and 18.0, which indicates some natural examples and OEs cannot be distinguished easily in the feature space. In Figure~\ref{fig:AT_f} (c) and Figure~\ref{fig:AT_f} (d), there is a skewed distribution between 0.85 and 0.99 for the cosine similarity of natural examples and AEs, and there is a skewed distribution between 0.5 and 2.5 for the $L_2$ distance, which indicates the feature of some natural examples and AEs is not adequately aligned. Thus, there is still large room for optimization of the feature of AT.

\emph{Q. 2 What characteristics does ideal robust representation have?}



Based on the observation, we propose two characteristics of robust feature: \textbf{Exclusion}: \emph{the feature of natural examples keeps away from that of other classes}; \textbf{Alignment}: \emph{the feature of natural and corresponding adversarial samples is close to each other}. First, Exclusion confirms the separability between different classes and avoids confusion in the feature space, which makes it hard to fool the model because the feature of different classes keeps a large distance. Second, Alignment ensures the feature of natural examples is aligned with adversarial ones, which guarantees the predicted results of the natural and adversarial examples of the same instances are also highly consistent. And it helps to narrow the gap between robust accuracy and clean accuracy. The two characteristics can serve as criteria for robust feature.

\emph{Q. 3 How to leverage the criteria of robust representation to improve AT?}


We further propose a generic AT framework with the \textbf{A}symmetric \textbf{N}egative \textbf{C}ontrast and \textbf{R}everse \textbf{A}ttention \textbf{(ANCRA)}, to concentrate on robust representation with the guidance of the two characteristics. Specifically, we suggest two strategies to meet the two criteria, respectively. For Exclusion, we propose Asymmetric Negative Contrast based on predicted probabilities (ANC), which freezes natural examples and pushes away OEs by reducing the confidence of the predicted class when predicted classes of natural examples and OEs are consistent. For Alignment, we use Reverse Attention (RA) to weight feature by parameters of the linear classifier corresponding to target classes, which contains the importance of feature to target classes during classification. Because the feature of the same class gets the same weighting and feature of different classes is weighted disparately, natural examples and AEs become close to each other in the feature space. Empirical evaluations show that existing methods combined with our framework can greatly enhance robustness, which implies the neglect of learning robust feature is one of the main reasons for the limited robust performance of AT. Our main contributions are summarized as follows:

\begin{itemize}
\item We find AT has flaws in the representation, and highlight Exclusion as well as Alignment as criteria for optimizing robust representation. 
\item We propose a generic defense framework, ANCRA, to obtain robust feature by the asymmetric negative contrast and reverse attention, with the guidance of the criteria for robust representation. It can be easily combined with other defense methods in a plug-and-play manner.
\item Empirical evaluations show our framework can obtain robust feature and greatly improve adversarial robustness, which achieves state-of-the-art performances on CIFAR-10, CIFAR-100 and Tiny-ImageNet. 
\end{itemize}

\section{Related Work}
\subsection{Adversarial Training} 

The mainstream view is that AT is the most effective defense, which has a training process of a two-sided game. The attacker crafts perturbation dynamically to generate adversarial data to cheat the defender, and the defender minimizes the loss function against adversarial samples to improve the robustness of models. It can be formalized as the min-max optimization problem:



Mardary~\etal\cite{PGD} propose PGD attack and PGD-based adversarial training, forcing the model to correctly classify adversarial samples within the epsilon sphere during training to obtain robustness, which is the pioneer of adversarial learning. Zhang~\etal\cite{TRADES} propose to learn both natural and adversarial samples and reduce the divergence of classification distribution of both to reduce the difference between robust accuracy and natural accuracy. Wang~\etal\cite{mart} find that misclassified samples during training harm robustness significantly, and propose to improve the model's attention to misclassification by adaptive weights. Zhang~\etal\cite{FAT} propose to replace fixed attack steps with attack steps that just cross the decision boundary, and improve the natural accuracy by appropriately reducing the number of attack iterations. Huang~\etal\cite{SAT} replace labels with soft labels predicted by the model and adaptively reduce the weight of misclassification loss to alleviate robust overfitting problem. Dong~\etal\cite{TE} also propose a similar idea of softening labels and explain the different effects of hard and soft labels on robustness by investigating the memory behavior of the model for random noisy labels. Chen~\etal\cite{SmoothandDistillation} propose random weight smoothing and self-training based on knowledge distillation, which greatly improves the natural and robust accuracy. Zhou~\etal\cite{MAN} embed a label transition matrix into models to infer natural labels from adversarial noise. However, little work has been done to improve AT from the perspective of robust feature learning. Our work shows AT indeed has defects in the feature distribution, and strategies proposed to learn robust feature can greatly advance robustness, which indicates the neglect of robust representation results in poor robust performance of AT.

\subsection{Adversarial Contrastive Learning}  

ACL is a kind of Contrast Learning (CL)~\cite{simclr,moco,BYOL} that extends to AT. Kim~\etal\cite{ROCL} propose to maximize and minimize the contrastive loss for training. Jiang~\etal~\cite{ACL_N} leverage a recent contrastive learning framework to maximize feature consistency, demonstrating that ACL pre-training can improve semi-supervised adversarial training. Xu~\etal~\cite{RCS} notice that ACL needs tremendous running time and propose a robustness-aware corest selection method to search for an informative subset. Fan~\etal\cite{ADVCL} notice that the robustness of ACL relies on fine-tuning, and pseudo labels and high-frequency information can advance robustness. Kucer~\etal\cite{ASSET} find that the direct combination of self-supervised learning and AT penalizes non-robust accuracy. Bui~\etal\cite{ASCL} propose some strategies to select positive and negative examples based on predicted classes and labels. Yu~\etal\cite{A-InfoNCE} find the instance-level identity confusion problem brought by positive contrast and address it by asymmetric methods. These methods motivate us to further consider how to obtain robust feature by contrast mechanism. We design a new negative contrast to push away natural and negative examples and mitigate the confusion caused by negative contrast.

\subsection{Robust Feature Learning} 

Robust feature learning is widely considered in various tasks. Although many AT approaches claim to aim for learning robust representation, their main focus often revolves around aligning the distributions of natural and adversarial feature, such as ALP/CLP~\cite{ALP}, TRADES~\cite{TRADES}, MMA~\cite{MMA}. There is a notable absence of in-depth discussions regarding the criteria and specifics of robust feature in their work. 

Yang~\etal~\cite{FACM} design FA and CMPD modules to collaboratively correct the feature retained in the intermediate layers and utilize the diversity among modules to improve robustness. Xiao~\etal\cite{KWTA} take the maximum k feature values in each activation layer to increase adversarial robustness. Zoran~\etal\cite{zoran2020} use a spatial attention mechanism to identify important regions of the feature map. Bai~\etal\cite{CAS} propose to suppress redundant feature channels and dynamically activate feature channels with the parameters of additional components, which build a linear layer to learn the importance of feature to target classes. Yan~\etal\cite{cifs} propose to amplify the top-k activated feature channels based on \cite{CAS}. Existing work has shown enlarging important feature channels is beneficial for robustness, but most approaches rely on extra model components and do not explain the reason. We propose the reverse attention to weight feature by class information without any extra components and explain it by Alignment of feature.

\begin{figure*}[!htbp]
    \vspace{-0.0em}
   \centering
   \includegraphics[width=1.0\textwidth]{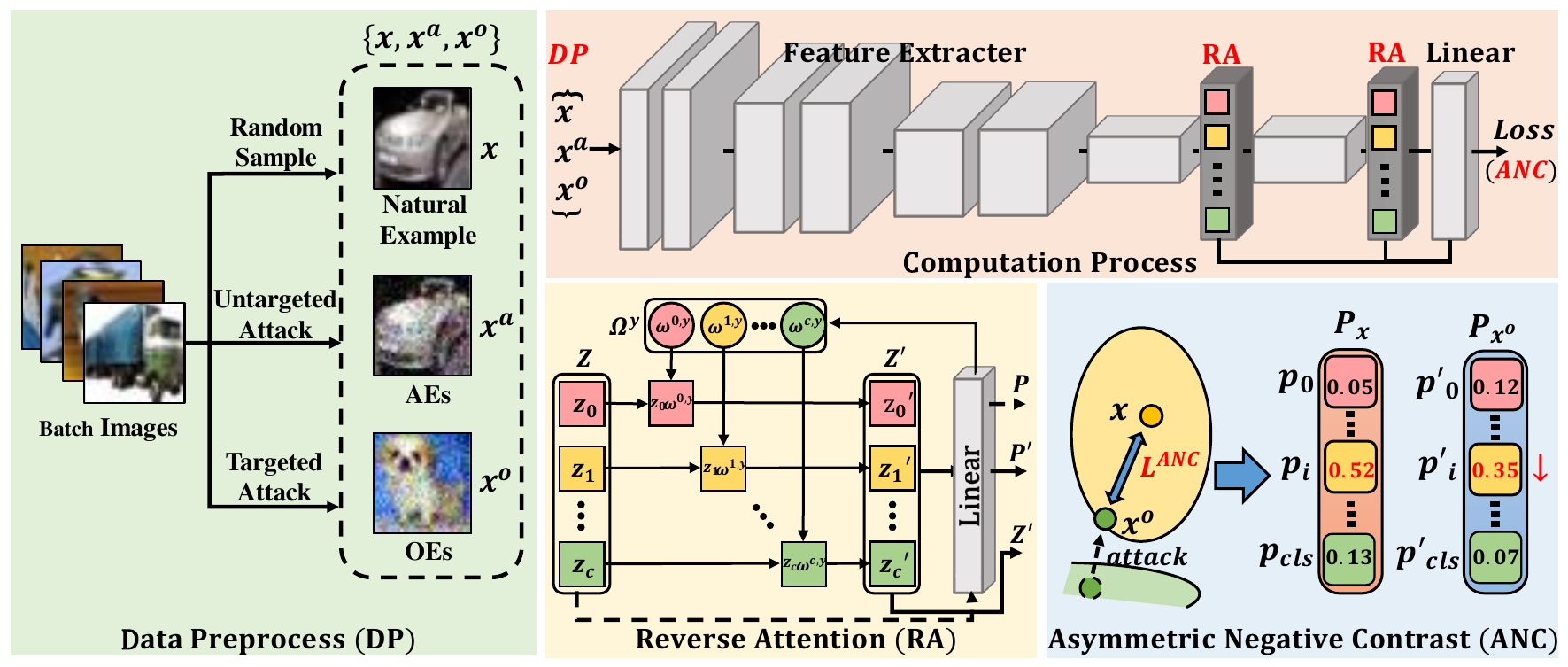}
   \caption{Overview of our proposed ANCRA. During DP (green), we sample a natural example $x$ with a label $y$ from the batch, generate an AE $x^{\alpha}$ by the untargeted attack, and craft an OE $x^o$ by the targeted attack aimed at $y$. RA (yellow) employs feature vectors and the parameters of the linear layer to calculate without extra parameters, and align feature of the same class by weighting them by the same subvector of the linear layer. ANC (blue) is a new loss term with a contrast mechanism. It introduces a repulsive force between $x$ and $x^o$ when the predicted classes of them are the same. To prevent the side effects of repulsion, it precisely decreases the classification probability of $x^o$ on the predicted class of $x$. Computation Process (peachpuff) shows the placement of DP, ANC and RA in the framework.}
   \label{fig:overview}
\end{figure*}

\section{Methodology}


This section details the instantiation of our AT framework, focusing on the two key criteria of robust feature. To meet Exclusion, we introduce an asymmetric negative contrast based on predicted probabilities to push away the feature of natural examples and OEs. Besides, we propose a strategy for generating negative examples through the targeted attack, leveraging prior knowledge of adversarial examples to enhance robustness. To confirm Alignment, we propose the reverse attention to weight the feature of the same class by specific weights, which are the corresponding parameters of the targeted class in the linear classifier. The weighting can minimize the feature gap between natural examples and Adversarial Examples without any extra modules. An overview of our framework is shown in Figure~\ref{fig:overview}.

\subsection{Notations}

 In this paper, capital letters indicate random variables or vectors, while lowercase letters represent their realizations. We define a model as $f(\cdot)$. $\Omega$ denotes the weights of the linear layer, which has chl (the number of feature channels) columns and cls (the number of classes) rows. Let $\mathcal{B}=\{x_i, y_i\}^{N}_{i}$ be a batch of natural samples. $x^a$ denotes \textbf{A}dversarial \textbf{E}xamples (AEs), and $x^o$ denotes the examples randomly selected from other classes (OEs) different from the label of $x$. Given our focus on feature pairs composed of $\{$natural samples, AEs$\}$ and $\{$natural samples, OEs$\}$, we refer to these feature pairs as PPs and NPs for the sake of brevity. For an input $x$, we define its feature as $Z$, the probability vector as $P$ and the predicted class as $h$, respectively.


\subsection{Adversarial Training with Asymmetric Negative Contrast}

\begin{figure}[!htbp]
\vspace{-0.0em}
    \centering
    \includegraphics[width=0.5\textwidth]{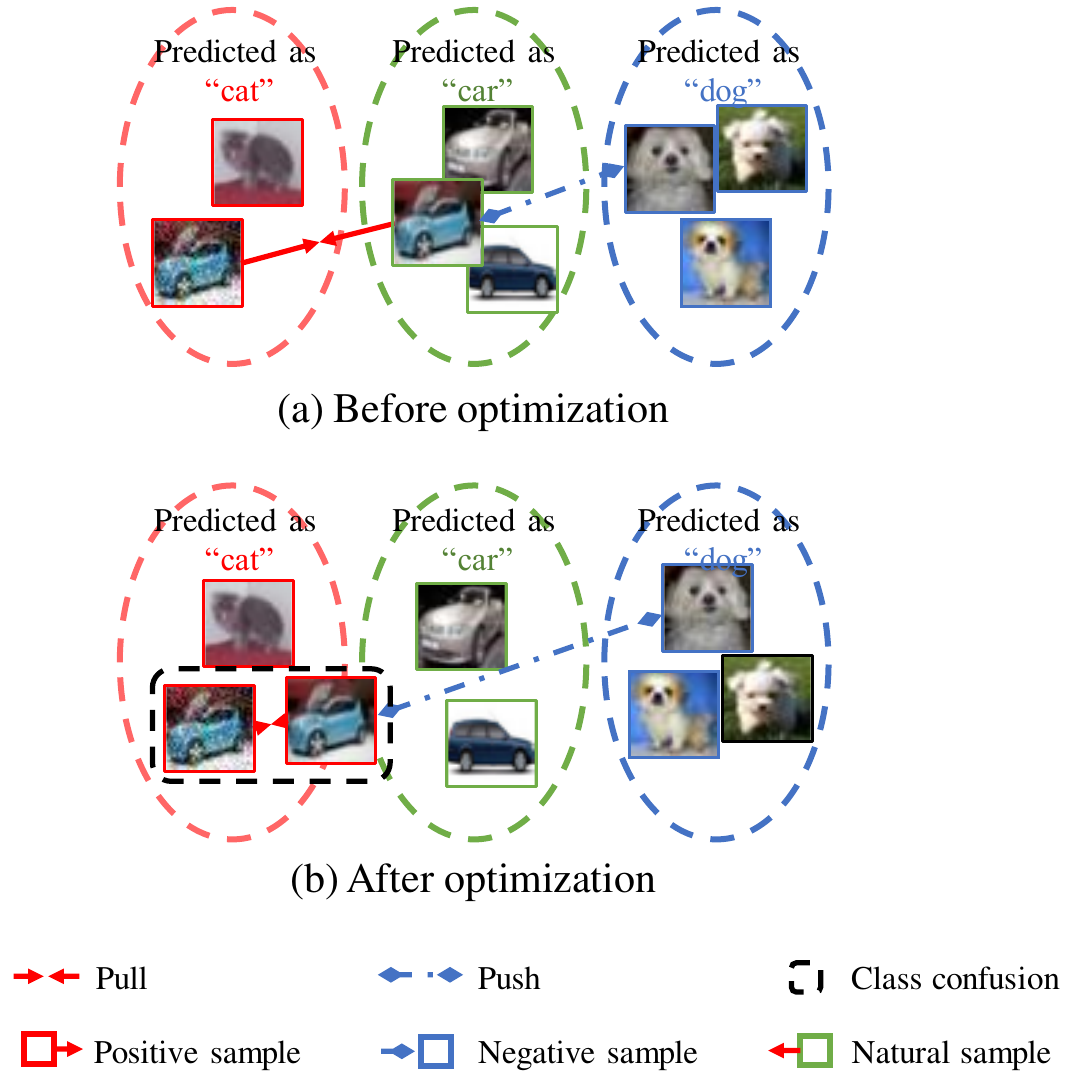}
    \caption{Illustrations of class confusion. In each circle, data points have the same predicted class. The negative contrast (blue lines) pushes natural examples to leave the initial class. Finally, natural examples come to the decision boundary and even into the wrong class easily as (b) shows.}
    \label{fig:identity_confusion}
\vspace{-0.0em}
\end{figure}

 First, we promote AT to learn robust representation that meets Exclusion. Notice that ACL has the contrastive loss~\cite{InfoNCE} to maximize the consistency of PPs and to minimize the consistency of NPs. Motivated by the contrast mechanism, we consider designing a new negative contrast and combining it with AT loss, which creates a repulsive action between NPs to keep large margins between different classes. 
\begin{equation}
\label{eq1}
\mathcal{L}^{NC}=\operatorname{Sim}\left(x, x^{o}; f(\cdot)\right),
\end{equation}
where $Sim$ is a similarity function, $x^o$ serves as the negative examples for $x$. However, \cite{A-InfoNCE} have indicated that when the predicted classes of the adversarial positive examples (i.e., AEs) and negative samples (i.e., OEs) are the same, there is a conflict led by the positive contrast, resulting in wrong classification. On this basis, we find a similar conflict can also be caused by the negative contrast when their predicted classes are different, called class confusion. We show a practical instance in Figure~\ref{fig:identity_confusion}. When optimizing the class space, the negative example pushes the natural example to leave the initial class. With this action, the training process suffers from class confusion, leading to natural examples moving toward the wrong class space, which does harm to Exclusion.

To alleviate the problem of class confusion, We should reasonably control the repulsion of negative contrast. We propose an asymmetric method of the negative contrast, $\mathcal{L'}^{NC}$, to decouple the repulsive force into two parts. It contains a one-side push from the natural example to the OE and a one-side push from the OE to the natural example, given by:
\begin{equation}
\label{eq2}
\mathcal{L'}^{NC}= \alpha \cdot \overline{\operatorname{Sim}}(Z, Z^o) + (1- \alpha) \cdot \overline{\operatorname{Sim}}(Z^o, Z),
\end{equation}
where $\overline{\operatorname{Sim}}(Z, Z^o)$ denotes the one-sided similarity of $x$ and $x^o$. When minimizing $\overline{\operatorname{Sim}}(Z, Z^o)$, we stop the back-propagation gradient of $Z$ and only move $Z^o$ away from $Z$. $\alpha$ denotes the weighting factor to adjust the magnitude of the two repulsive forces. When $\alpha$ = 0, negative samples are frozen and only the feature of natural samples is pushed far away from the feature of negative samples. As $\alpha$ increases, the natural sample becomes more repulsive to the negative sample and the negative sample pushes the natural example less. To mitigate the class confusion problem, we should choose $\alpha$ that tends to 1 to reduce the repulsive force from the negative sample to the natural example, to prevent the natural example from being pushed into the wrong class. 

Moreover, we propose the negative contrast based on predicted probabilities, $\mathcal{L}^{ANC}$, to measure the repulsive force of NPs pushing away from each other. It pushes away NPs by decreasing the corresponding probabilities of the predicted classes when the predicted classes of NPs are consistent, as shown in Asymmetric Negative Contrast in Figure~\ref{fig:overview}.
\begin{equation}
\label{eq3}
\mathcal{L}^{ANC}= \mathbb{I}\left(h=h^o\right) \cdot \left[ \alpha \sqrt{\hat{p_{h}} p^o_{h}} + (1 - \alpha) \sqrt{p_{h} \hat{p^o_{h}}} \right],
\end{equation}
where $\mathbb{I}(\cdot)$ denotes the Indicator function and $\hat{p}$ denotes freezing the back-propagation gradient of $p$. $h$ and $h^o$ denote the predicted classes of the NP. And $p_h$ and $p^{o}_{h}$ denote the predicted probabilities of class $h$ of the NP. Under the negative contrast, the model pushes the natural example in the direction away from the predicted class of the OE, and pushes the OE away from the predicted class of the natural example when and only when two predicted classes of the NP are consistent. This ensures that the action of Exclusion not only pushes away the feature of NPs in the feature space, but also reduces the probabilities of NPs in the incorrect class. Since the negative contrast has only directions to reduce the confidence and no explicit directions to increase the confidence, it does not create any actions to push the natural example into the feature space of wrong classes even in the scenario of class confusion, which can effectively alleviate the problem. Now we can combine it with an AT loss to have new loss function:
\begin{equation}
\label{eq4}
\mathcal{L}oss=\mathcal{L}^{AT} + \zeta \cdot \mathcal{L}^{ANC}, 
\end{equation}
where $\zeta$ is the weight of $Sim$. When minimizing Equation~\ref{eq4}, $\mathcal{L}^{\mathrm{AT}}$ learns to classify natural examples and AEs correctly, and additional negative contrast $\mathcal{L}^{ANC}$ promotes the inconsistency of NPs, which keeps the feature of NPs away from each other to ensure Exclusion.

\subsection{Negative Samples Generation by Targeted Attack}


Due to negative contrast, we need to obtain appropriate OEs as negatives to optimize the representation, which is shown in the Data Preprocess in Figure~\ref{fig:overview}. Previous negative sampling strategies~\cite{FNC} (Soft-LS, Hard-LS, Random) simply screen natural samples and pick up the negatives from them, but rarely consider generating special negative samples to assist learning. We innovatively design a strategy to craft OEs (Targeted) by the targeted attack: natural negative examples with labels that are different from those of natural examples are attacked to the labeled classes of natural examples. We have chosen the targeted PGD-10~\cite{PGD} in the experiment, given by:
\begin{equation}
\label{eq5}
x_{t+1}^{o}:=\mathop{\boldsymbol{\Pi}} \limits_{\mathbb{N}(x^o, \epsilon)}\left(x_{t}^{o} - \epsilon \operatorname{sign}\left(\nabla_{x^o} \mathcal{L}_{CE}\left(\left(f(x_{t}^{o}\right), y\right)\right)\right),
\end{equation}
where $\mathbb{N}(x^o, \epsilon)$ represents $\{\tilde{x}:\|\tilde{x}-x^o\|_{\infty} \leq \epsilon\}$, $\epsilon$ is the perturbation budget. $x_{t}^{a}$ denotes adversarial samples after the $t$th attack iteration, $\boldsymbol{\Pi}$ denotes a clamp function from 0 to 1, $sign$ denotes a sign function, $\mathcal{L}_{CE}$ denotes the cross-entropy loss and $\nabla_{x^o} \mathcal{L}_{CE}$ denotes the gradient of $\mathcal{L}_{CE}$ with respect to $x^o$. Details of all the strategies are shown in Table~\ref{tbstrategies}.

\begin{table}[!htbp]
\centering
\caption{Strategies of negative samples. $h(\cdot)$ denotes the predicted class, $x$ denotes a natural example in the current batch, $y$ denotes the ground-truth label of $x$, respectively.}

\begin{tabular}{ll}

\toprule

Strategy&Condition\\

\hline

Random&$\{ x_i \vert  x_i \in \mathcal{B}, y_i \neq y\}$ \\

Soft-LS&$\{ x_i \vert  x_i \in \mathcal{B}, y_i \neq y, h(x_i) = h(x)\}$ \\

Soft-LS&$\{ x_i \vert  x_i \in \mathcal{B}, y_i \neq y, h(x_i) = y\}$ \\

Targeted& $\{ {x'}_i \vert  x_i \in \mathcal{B}, y_i \neq y, {x’}_i = \mathop{max} \limits_{\mathbb{N}(x_i, \epsilon)} \mathcal{L}_{CE}\left(f( \tilde{x_i} ), y\right)\}$\\

\bottomrule
\end{tabular}
\label{tbstrategies}
\end{table}


 The motivation makes intuitive sense. (1) The negative adversarial sample generated by the targeted attack will be classified as the labeled class of the natural example with high confidence, which makes it a very hard negative sample. (2) The negative adversarial sample contains adversarial noise, which is special feature that natural negative samples do not have. This feature helps the model learn the paradigm of adversarial noise and improve its robust performance. Besides, it has an extra advantage. (3) When the number of classes or batch size of the dataset is small, the three sampling methods often fail ($\geq 60\%$ on CIFAR-10 with a batch size of 128). In contrast, our approach of generating OEs through the targeted attack is guaranteed not to fail.


\subsection{Adversarial Training with Reverse Attention}


Second, we continue to improve AT to learn robust representation that meets Alignment. Motivated by~\cite{CAS, cifs}, we utilize the values of linear weight to denote the importance of feature channels to targeted classes. We exploit the importance of feature channels to align the examples in the same classes and pull close the feature of PPs, which is named by reverse attention. To be specific, we take the Hadamard product of the partial parameters of the classifier $\Omega^j$ and the feature vector $Z$. “partial parameters” mean those weights of the linear layer that are used to calculate the probability of the target class. Because the reverse attention weights the feature of PPs by the same parameters, it helps Alignment. Given by:

\begin{equation}
 \quad {z'}_i= \begin{cases}z_i \cdot \omega^{i,y}, & \text { (training phase) } \\ z_i \cdot \omega^{i,h(x)}, & \text { (testing phase) }\end{cases}
 \label{eq6}
\end{equation}


where $z_i$ denotes the $i$th feature channel of the feature vector, $\omega^{i,j}$ denotes the linear parameters of the $i$th feature channel to the $j$th class. We utilize the weights of the linear layer to represent the importance of features for classification, and directly apply this information to the feature vector through element-wise multiplication. During the training phase, we use the true label $y$ as an indicator to determine the importance of channels. In the testing phase, since the true label is not available, we simply choose a sub-vector of the linear weight by the predicted class $h(x)$ as the importance of channels. The model with the reverse attention does not need any extra modules, but module interactions are changed. 


We add reverse attention in the final layer to ensure that features are fused with prior knowledge from feature to classes before being processed by the linear classifier. As shown in Reverse Attention in Figure~\ref{fig:overview}, the computation flow produces the weighted feature vectors $Z'$ by the dot product and offers a new predicted probability vector $P'$ from $Z'$. Simultaneously, we train the linear layer with the unweighted feature to ensure the weights of the linear layer consistently contain information of feature importance throughout the training process. Thus we calculate the auxiliary loss with $P$. This approach has two advantages: (1) The feature vector incorporates prior information from feature to categories in an attention-mechanism manner, making itself more conducive to correct classification. (2) Feature vectors of the same label become more similar, while feature vectors of different classes become more distinct, facilitating correct classification.


Let's make a detailed analysis and explanation of the principle of this method. In the model, the feature extractor captures the representation that contains enough information to classify, and the linear layer establishes a relationship from feature to predicted classes. The probability of the predicted class equals the sum of the product of linear weight corresponding to predicted class and feature vector. In this premise, the linear layer learns to correctly increase the probability of the label class and decrease other probabilities when training. Thus it can gradually recognize which feature channels are important for specific classes, and keep large weight values for those feature channels. On this basis, we propose reverse attention to utilize its parameters containing feature importance to improve feature. (1) From the perspective of the feature itself, \textbf{the feature vectors are multiplied by the parameters of the target class, which can change the magnitude of each feature channel adaptively according to the feature importance}, acting as attention with the guidance of the linear layer. The important channels in the feature vector are boosted and the redundant channels are weakened after the attention. Therefore, the feature value contributing to the target class will become larger, which is helpful for correct classification. (2) From the perspective of the overall feature distribution, reverse attention can induce beneficial changes in the feature distribution. \textbf{Since the linear layer is unique in the model, different examples in the same class share the same linear weights}. Feature vectors with the same target class(e.g., examples in PPs) get the same weighting and become more similar. Moreover, feature vectors with different target classes(e.g., examples in NPs) are weighted by different parameters, and the weighted feature distributions may become more inconsistent. Therefore, the reverse attention guides the alignment of the feature of the examples in the same class, pulling the feature of PPs closer and pushing the feature of NPs far away, which benefits Alignment and drops by to promote Exclusion. The aligned feature has similar activations in every feature channel, which helps the model narrow the gap between feature of natural examples and AEs.

\subsection{Algorithm of ANCRA}

We show the whole computation of our ANCRA as follows: 

\begin{algorithm}
\caption{Asymmetric Negative Contrast and Reverse Attention(ANCRA)}\label{alg:ra}
\begin{algorithmic}
\REQUIRE Training dataset $\mathcal{D}$, the model $f(\cdot)$, weight of loss $\gamma$, attacker $A$ (PGD-10), batch size $N$, number of epochs $T$.
\ENSURE the robust model $f(\cdot)$.
\FOR{t=1 to T}
    \FOR{mini-batch $\mathcal{B}=\{(x_i, y_i)\}^{N}_{i} \subset \mathcal{D}$}
        \STATE $\mathcal{L} \gets 0$
        \FOR{i=1 to N}
            \STATE $x, y \gets x_i, y_i$ 
            \STATE $x^a \gets A(x)$;
            \STATE Generate $x^o$ by Equation~\ref{eq5};
            \STATE $\widetilde{x} \gets \{x, x^a, x^o\}$;
            \STATE $Z \gets f^{0: -4}(\widetilde{x})$;
            \FOR{j=-3 to -2}
                \STATE $Z \gets f^j(Z)$;
                \STATE $P \gets f^{-1}(Z)$;\enspace//Get auxiliary probability vector
                \STATE Compute $\mathcal{L}oss_j$ with $P$ by Equation~\ref{eq4};
                \STATE Compute $Z'$ as Equation~\ref{eq6} ;  
                \STATE $Z \gets Z'$; 
            \ENDFOR
            \STATE $P' \gets f^{-1}(Z')$;\enspace //Get the final probability vector
            \STATE Compute $\mathcal{L}oss$ with $P'$ by Equation~\ref{eq4};
            \FOR{j=-2 to -1}
                \STATE $\mathcal{L}oss \gets \mathcal{L}oss + \gamma \cdot \mathcal{L}oss_j$;
            \ENDFOR
        \ENDFOR
        \STATE $\mathcal{L} \gets \mathcal{L} + \frac{1}{N}\mathcal{L}oss$;
    \ENDFOR
    \STATE Backpropagation and optimize $f(\cdot)$;
\ENDFOR
\RETURN $f(\cdot)$
\end{algorithmic}
\end{algorithm}

where $f^{j}(\cdot)$ denotes the $j$th module of the model, $f^{i:j}(\cdot)$ denotes the model components from the $i$th to $j$th layer. So $f^{-1}(\cdot)$ is the linear classifier, $f^{0:-2}(\cdot)$ represents the feature extracter. Notice that we calculate $\mathcal{L}oss_j$ in the RA layer to make sure the parameters used for weighting contain the correct information from feature to classes. When the shape of $Z$ is not suitable for the linear layer, we apply broadcast and reshape to fit it.

\section{Experiments}

In order to demonstrate the effectiveness of the proposed approach, we show feature distribution and visualization of trained models first. Then we evaluate our framework against white-box attacks, adaptive attacks and black-box attacks to make a comparison with other defense methods. We conduct experiments across different datasets and models. We further make evaluations in the deployment scenario. Because our methods are compatible with existing AT techniques and can be easily incorporated in a plug-and-play manner, we choose three baselines~\cite{PGD, TRADES, mart} to combine with our framework for evaluation: PGD-AT-ANCRA, TRADES-ANCRA, and MART-ANCRA.

\subsection{Settings}


\textbf{Implementation}
We train ResNet~\cite{ResNet} and WideResNet~\cite{WideResNet} models on CIFAR-10 and CIFAR-100~\cite{CIFAR-10}, and PreActResNet~\cite{preactresnet} models on Tiny-ImageNet~\cite{TinyImageNet}. CIFAR-10 dataset contains 60,000 color images having a size of $32\times32$ in 10 classes, with 50,000 training and 10,000 test images. CIFAR-100 dataset contains 50,000 training and 10,000 test  images in 100 classes. We adopt the SGD optimizer with \textbf{a learning rate of 0.01}, a weight decay of ${2.0\times10^{-4}}$, epochs of 120 and a batch size of 128 as~\cite{mart}. For the trade-off hyperparameters $\beta$, we use 6.0 in TRADES\footnote{Unlike vanilla TRADES, we maximize the CE loss to generate adversarial examples as PGD-AT and MART.} and 5.0 in MART. For other hyperparameters, we have tuned the values based on TRADES-ANCRA and set $\alpha=1.0, \zeta=3.0$. We set $\gamma=2$ as \cite{CAS}. We generate AEs for training by $L_{\infty}$-norm PGD~\cite{PGD}, with a step size of 0.007, an attack iterations of 10 and a perturbation budget of 8/255. We use a single NVIDIA A100 and two GTX 2080 Ti.


\textbf{Baseline}
We compare our ANCRA with the popular baselines: PGD-AT~\cite{PGD}, TRADES~\cite{TRADES}, MART~\cite{mart} and SAT~\cite{SAT}. Moreover, we also choose three state-of-the-art methods: AWP~\cite{AWP}, S2O~\cite{S2O} and FairARD~\cite{SAT_r}. For the fairness of comparison, we keep the same settings among all the baselines with our settings. Because some defense methods under our settings perform poorly, we extracted their best results from the paper, which are marked with $\dagger$.

\textbf{Evaluation}
Feature distribution and feature visualization map are chosen to show our superiority of robust representation learning. Besides, clean accuracy and robust accuracy are used as the evaluation metrics. We choose PGD~\cite{PGD}, FGSM~\cite{FGSM}, C\&W~\cite{C&W}, MIM~\cite{MIM} and AutoAttack~\cite{AA} to attack models. AutoAttack is an adaptive and reliable attack composed of three white-box attacks and one black-box attack. We notice that our methods use the auxiliary probability vector $P$ in the training phase, so we design customized adaptive attacks to maximize the total loss. Moreover, we attempt to evaluate RA in the deployment phase. There are four attack scenarios: (1) evaluation against white-box attacks, (2) evaluation against customized adaptive attacks, (3) evaluation against black-box attacks, and (4) evaluation in the deployment phase. The perturbation budget is 8/255 under the $L_{\infty}$. The attack iterations of PGD and C\&W are 40, and the step size of FGSM is 8/255 unlike 0.007 for other attacks.

\begin{figure*}[!htbp]

\vspace{-0.0em}
    \centering
    \includegraphics[width=1.0\textwidth]{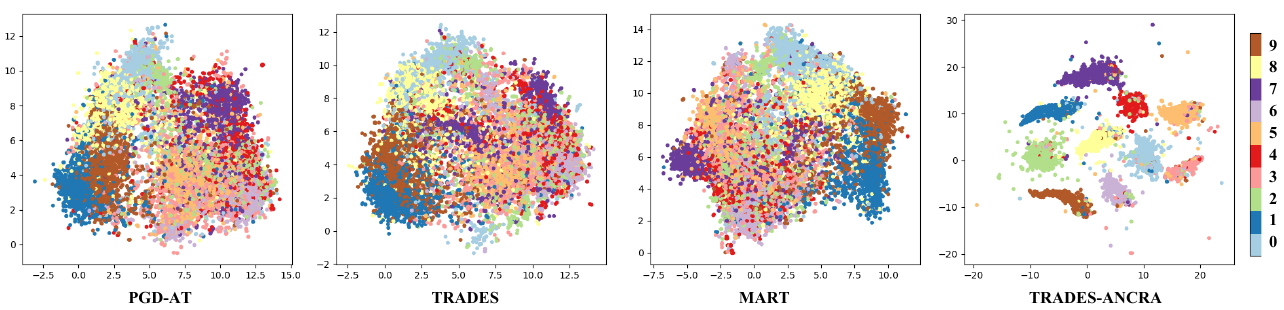}
    \caption{Feature visualization of four methods on natural and adversarial examples. Adversarial samples are crafted by PGD-10.}
    \label{fig:all}
\end{figure*}

\begin{figure*}[!htbp]
\vspace{-0.0em}
    \centering
    \includegraphics[width=1.0\textwidth]{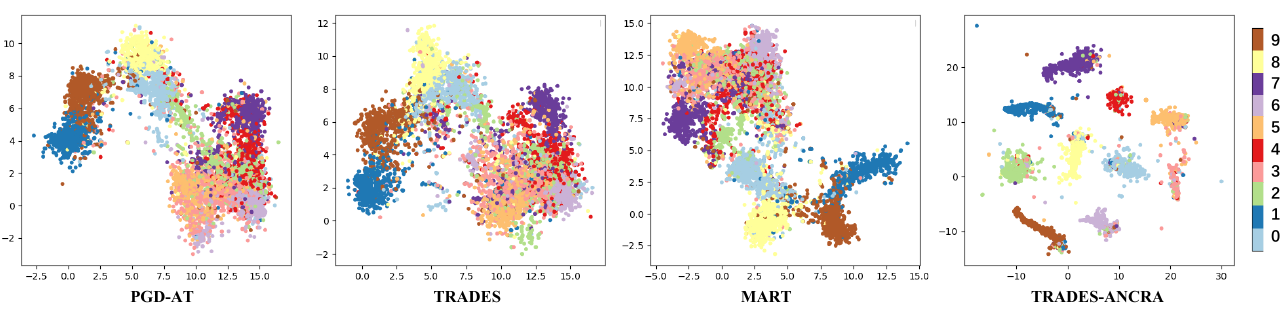}
    \caption{Feature visualization of four methods on natural examples.}
    \label{fig:nat}
\end{figure*}

\subsection{Feature Distribution and Visualization}

\begin{figure}[!htbp]
    \centering
    \includegraphics[width=0.5\textwidth]{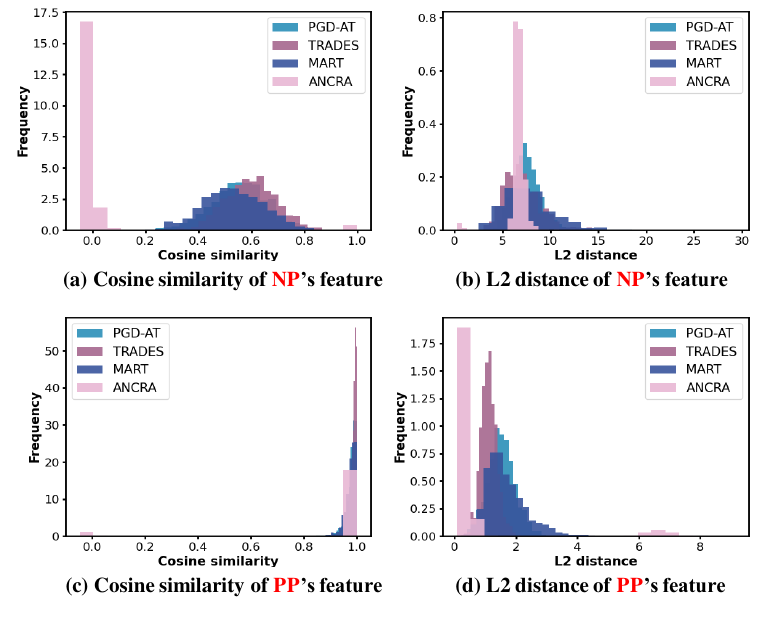}
    \caption{Frequency histograms of the {$L_2$} distance and cosine similarity of feature of natural examples, AEs and OEs. As shown in (a) and (b), the feature of NPs is adequately distant, satisfying Exclusion. As shown in (c) and (d), the feature of PPs is sufficiently close, meeting Alignment.}
    \label{fig:AT_all}
\end{figure}

Frequency histograms of feature distribution are shown in Figure~\ref{fig:AT_all}. In Figure~\ref{fig:AT_all} (a) and Figure~\ref{fig:AT_all} (b), the cosine similarity of our method between NPs shows a skewed distribution between -0.05 and 0.1, and the $L_2$ distance of our method shows a bell-shaped distribution between 5.5 and 10.0, which indicates NPs have been fully distinguished in the feature space and Exclusion has been met. As shown in Figure~\ref{fig:AT_all} (c) and Figure~\ref{fig:AT_all} (d), with our method there is a uniform distribution between 0.95 and 0.99 for the cosine similarity of the feature between PPs, and a skewed distribution between 0.05 and 1.5 for the $L_2$ distance, which indicates the feature between PPs is very close to each other and Alignment has been confirmed. It shows that our methods can greatly improve feature distribution, which follows the criteria of Exclusion and Alignment.

Besides, we use UMAP~\cite{UMAP}, a visualization technique, to draw the feature distribution map. Results are shown in Figure~\ref{fig:nat} and Figure~\ref{fig:all}, where different colors denote samples of different classes. It shows existing AT methods can learn good representations of natural examples but have been confused when dealing with natural and adversarial examples at the same time. This demonstrates their representations are not robust. Unlike traditional AT methods, our method can effectively discriminate between samples of different classes, including both natural and adversarial samples. The adversarial samples gather around the class centers rather than the boundaries. These indicate our framework successfully helps AT to obtain robust feature.

\subsection{Comparisons with Existing Defenses}

\begin{table*}[!htbp]
\small

\centering
\caption{Robustness (\%) against white-box attacks. Nat denotes clean accuracy. AA denotes robust accuracy against AutoAttack. AVG denotes average robust accuracy. The variation of accuracy $\leq 1.3\%$. We show the best results with \textbf{bold}. }
\begin{tabular}{ccccccccccccc}
\toprule

\multicolumn{1}{c}{\multirow{2}*{Defense}}& \multicolumn{6}{c}{CIFAR-10}& \multicolumn{6}{c}{CIFAR-100} \\
\cmidrule(r){2-7} \cmidrule(r){8-13}
\multicolumn{1}{c}{}&Nat&PGD&FGSM&C\&W&AA&AVG\;&Nat&PGD&FGSM&C\&W&AA&AVG\\
\hline

PGD-AT&80.90&44.35&58.41&46.72&42.14&47.91\;&56.21&19.41&30.00&41.76&17.76&27.23\\

TRADES&78.92&48.40&59.60&47.59&45.44&50.26\;&53.46&25.37&32.97&43.59&21.35&30.82\\

MART&79.03&48.90&60.86&45.92&43.88&49.89\;&53.26&25.06&33.35&38.07&21.04&29.38\\

SAT&63.28&43.57&50.13&47.47&39.72&45.22\;&42.55&23.30&28.36&41.03&18.73&27.86\\

AWP&76.38&48.88&57.47&48.22&44.65&49.81\;&54.53&27.35&34.47&44.91&21.98&31.18\\

S2O&40.09&24.05&29.76&47.00&44.00&36.20\;&26.66&13.11&16.83&43.00&21.00&23.49\\



\hline

MART$\dagger$&83.07&55.57&65.65&54.87&$\backslash$&58.70\;&$\backslash$&$\backslash$&$\backslash$&$\backslash$&$\backslash$&$\backslash$\\

SAT$\dagger$&84.27&49.11&56.81&48.58&46.13&50.16\;&57.81&24.07&29.09&23.69&21.80&24.66\\

AWP$\dagger$&81.20&51.60&55.30&48.00&46.90&50.45\;&$\backslash$&$\backslash$&$\backslash$&$\backslash$&$\backslash$&$\backslash$\\

S2O$\dagger$&83.65&55.11&$\backslash$&$\backslash$&48.30&51.71\;&58.45&30.58&$\backslash$&$\backslash$&$\backslash$&30.58\\

FairARD$\dagger$&82.96&52.05&57.69&50.69&49.13&52.39\;&57.08&29.38&32.87&26.92&25.55&28.68\\

\hline\hline

PGD-AT-ANCRA&\pmb{85.10}&\pmb{89.03}&87.00&\pmb{89.23}&59.15&81.10\;&59.73&58.10&58.45&58.58&34.44&52.39\\

TRADES-ANCRA&81.70&\pmb{82.96}&82.74&83.01&\pmb{59.70}&77.10\;&53.73&51.24&52.17&52.55&\pmb{35.81}&47.94\\

MART-ANCRA&84.88&88.56&\pmb{87.95}&88.77&59.62&\pmb{81.23}\;&\pmb{60.10}&\pmb{58.40}&\pmb{58.74}&\pmb{59.41}&35.05&\pmb{52.90}\\

\bottomrule
\end{tabular}

\label{tb:white-box}
\end{table*}





\begin{table}[!htbp]
\centering
\caption{Clean and robust accuracy (\%) on Tiny-ImageNet.}
\begin{tabular}{ccccccc}
\toprule
\multicolumn{1}{c}{Defense}&Nat&PGD\\
\hline
PGD-AT&41.31&10.28\\

TRADES&37.27&16.30\\

MART&38.61&14.78\\

PGD-AT-ANCRA&43.02&29.79\\

TRADES-ANCRA&38.94&31.27\\

MART-ANCRA&43.83&31.44\\
\bottomrule
\end{tabular}
\label{tb:tiny}
\end{table}

\begin{table}[!htbp]
\centering
\caption{Comparative experiments with methods in the RobustBench. All the models are in ResNet-18 trained on CIFAR-10. AA denotes robust accuracy against AutoAttack. The best results are in \textbf{bold}.}

\begin{tabular}{ccc}

\toprule

\multicolumn{1}{c}{Defense}&Nat&AA\\

\hline
Sehwag~\etal~\cite{1}&\pmb{87.35}&58.50\\

Addepalli~\etal~\cite{2}&85.71&52.48\\

Addepalli~\etal~\cite{3}&80.24&51.06\\

\hline\hline

PGD-AT-ANCRA&85.10&59.15\\

TRADES-ANCRA&81.70&\pmb{59.70}\\

MART-ANCRA&84.88&59.62\\

\bottomrule
\end{tabular}

\label{tb:robustbench}
\end{table}

\textbf{Comparison results against white-box attacks}
 We have conducted experiments on ResNet-18 to evaluate different defenses under white-box attacks. The results are shown in Table~\ref{tb:white-box}. Since some defenses show terrible performance under our training setting, we excerpt several results reported in papers as a reference, marked with $\dagger$. First, on CIFAR-10, our approaches increase the clean accuracies by 5.2\%, 3.2\% and 5.9\% compared with based approaches, and also improve the robust performance under all the attacks (e.g., increase by 44.7\%, 34.6\% and 39.7\% against PGD). Compared with state-of-the-art defenses, the robust accuracies of our methods are almost two times as large as theirs (e.g., 81.23\% > 52.39\%). Second, on CIFAR-100, our approaches also greatly improve the robustness and advance the clean accuracies. The clean accuracies of our methods have been increased by 3.5\%, 0.3\% and 6.8\% compared with based methods, and the lowest average robust accuracy of ours is larger than the best one among other methods by 10.26\%. We also train PreActResNet-18 models on Tiny-ImageNet. As shown in Table~\ref{tb:tiny}, our methods made obvious progress in robustness and generalization compared with baselines. To our surprise, MART-ANCRA and PGD-ANCRA rather than TRADES-ANCRA gain the best performance in a lot of cases without hyper-parameter tuning.

  We have made a comparison with the current state-of-the-art performances listed on the RobustBench\footnote{\url{https://robustbench.github.io/}} on ResNet-18. The results are shown in Table~\ref{tb:robustbench}. Compared with those methods without synthetic or extra data (i.e., \cite{2} and \cite{3}), our method has a higher robust accuracy than theirs by 7.0\%. And our method has even outperformed the methods with synthetic data \cite{1} in robustness. Though the clean accuracy of \cite{1} is more than ours by 5.6\%-2.2\%, the best robust performance has indicated the effectiveness of our methods. Experiment results in ResNet-18 have shown our superiority of robustness.



\textbf{Comparison results against adaptive attacks}
We train ResNet-18 models on CIFAR-10 and CIFAR-100, and we also train WideResNet-28-10 and WideResNet-34-10 on CIFAR-10. Besides, we report the results of vanilla TRADES as a baseline. We report the performance against customized adaptive attacks with $P$ to evaluate the robustness. As shown in Table~\ref{tab:adaptive} and Table~\ref{Tab:Wide}, the robust accuracies of our method against adaptive attacks are larger than those of the baseline against vanilla attacks. For example, robustness on ResNet-18 against adaptive PGD is higher than the baseline by 13.28\% and robustness on WideResNet-34-10 against adaptive PGD is higher than the baseline by 2.88\%. The robustness under adaptive AutoAttack has increased slightly, but not by a significant magnitude (0.74\%, 1.20\%). We will discuss the reasons in the Limitation. The results indicate that our approaches can still maintain superb performance under adaptive attacks. 


\textbf{Comparison results against black-box attacks}
We have made some experiments against transfer-based black-box attacks on ResNet-18. Notice that all the models are ResNet-18, so adversarial examples are easy to be transfered. AEs are generated by PGD-100 and MIM-100 on source models and tested on target models.  As shown in Table~\ref{Tab:black}, the robustness agaisnt MIM-100 of TRADES-ANCRA is better than that of baseline methods by 17.30\%, 3.12\% and 3.37\%. And the robust accuracies of ours against PGD-100 are also higher than those of baseline approaches. It shows our method gains the best black-box robustness among all the methods, indicating its effectiveness in the black-box scenario.

\begin{table}[!htbp]
\setlength\tabcolsep{5.0pt}
\centering
\caption{Robustness (\%) against adaptive attacks.}
\label{tab:adaptive}
\begin{tabular}{ccccccc}
\toprule
\multicolumn{1}{c}{\multirow{2}*{Defense}}& \multicolumn{3}{c}{CIFAR-10}& \multicolumn{3}{c}{CIFAR-100}\\
\cmidrule(r){2-7} 
\multicolumn{1}{c}{}&PGD&FGSM&C\&W&PGD&FGSM&C\&W\\
\hline
TRADES&48.40&59.60&47.59&25.37&32.97&43.59\\

TRADES-ANCRA&61.68&61.56&72.36&31.68&33.03&43.91\\



PGD-AT-ANCRA&54.43&58.23&66.36&26.07&32.42&43.10\\



MART-ANCRA&56.96&60.43&71.06&28.54&33.12&43.25\\


\bottomrule
\end{tabular}
\end{table}

\begin{table}[!htbp]
\centering
\caption{Robust accuracy (\%) against adaptive attacks on WideResNet (WRN) models.}
\label{Tab:Wide}

\begin{tabular}{ccccccc}
\toprule

\multicolumn{1}{c}{\multirow{2}*{Defense}}& \multicolumn{1}{c}{\multirow{2}*{Model}}& 
\multicolumn{2}{c}{Adaptive Attacks}\\

\cmidrule(r){3-4} 
\multicolumn{1}{c}{}&\multicolumn{1}{c}{}&\multicolumn{1}{c}{PGD}& \multicolumn{1}{c}{AA}\\

\hline
TRADES&WRN-28-10&57.08&51.11\\

TRADES-ANCRA&WRN-28-10&58.60&51.85\\

TRADES&WRN-34-10&56.47&50.79\\

TRADES-ANCRA&WRN-34-10&59.35&51.99\\

\bottomrule
\end{tabular}

\end{table}

\begin{table}[!htbp]
\setlength\tabcolsep{7.0pt}
\centering
 \caption{Robustness (\%) against transfer-based attacks. }
\begin{tabular}{ccccc}
\toprule

\multicolumn{1}{c}{\multirow{2}*{Attack}}&\multicolumn{1}{c}{\multirow{2}*{Target}}& \multicolumn{3}{c}{Source}\\
\cmidrule(r){3-5} 
\multicolumn{2}{c}{}&PGD-AT&TRADES&MART\\

\hline
\multicolumn{1}{c}{\multirow{4}*{PGD-100}}&PGD-AT&44.28&58.37&59.67\\

\multicolumn{1}{c}{}&TRADES&58.96&48.33&60.33\\

\multicolumn{1}{c}{}&MART&58.69&58.59&48.86\\

\multicolumn{1}{c}{}&TRADES-ANCRA&62.28&60.66&62.46\\

\hline
\multicolumn{1}{c}{\multirow{4}*{MIM-100}}&PGD-AT&44.73&58.25&59.65\\

\multicolumn{1}{c}{}&TRADES&58.91&48.53&60.21\\

\multicolumn{1}{c}{}&MART&58.66&58.46&49.26\\

\multicolumn{1}{c}{}&TRADES-ANCRA&62.03&60.43&62.23\\

\bottomrule
\end{tabular}
 \label{Tab:black}
\end{table}

\subsection{Defense Results in Deployment}

\begin{figure}[!htbp]
    \centering
    \includegraphics[width=0.5\textwidth]{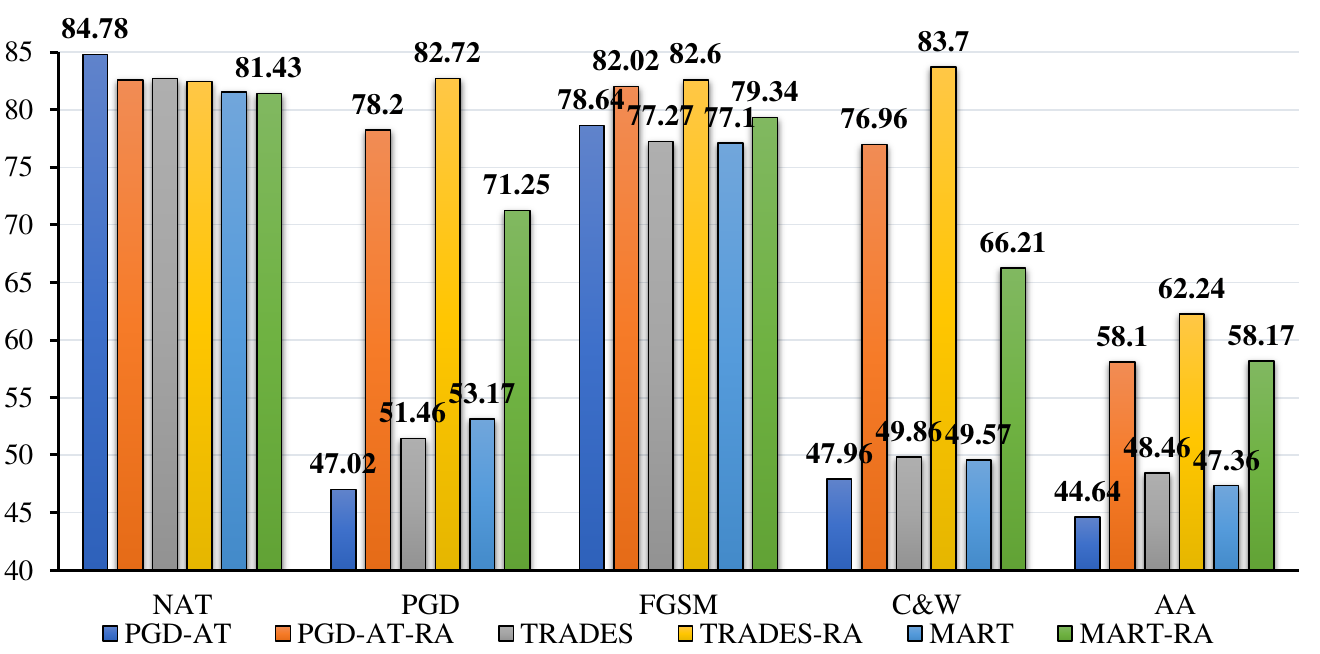}
    \caption{Defense results during the deployment phase. We take model parameters trained by PGD-AT, TRADES, MART and load them on standard ResNet-18 models and models with RA. And then we test them with four attacks. Each model with RA has a great enhancement of robustness at the cost of marginal reduction of clean accuracy.}
    \label{fig:deploy}
\end{figure}

Suppose the attackers illicitly obtain permissions of users with the access to input, output, and gradient information, yet the model's structure remains undisclosed. The attackers may craft adversarial examples to disrupt the normal operation of the model. This is a common scenario if people use AI agent models and online APIs of Large Language Models (e.g., GPT-4, LLaMA, Grok). How can we handle these adversarial examples? Notice that our RA is a parameterless method, we wonder if RA can also work in the deployment phase. We add an RA layer to the standard model and load the model parameters trained by other AT methods on it. We select models trained by PGD-AT, TRADES and MART on ResNet-18 and evaluate them by white-box attacks. As shown in Figure~\ref{fig:deploy}, the robustness against PGD-40 of PGD-AT, TRADES and MART has increased by 31.18\%, 31.26\% and 18.08\%. All the robust accuracies have greatly enhanced with a marginal decrease of clean accuracies, which indicates RA can improve robustness in the deployment phase.

\subsection{Ablation Studies} 

\textbf{Analysis of Components} We train four models by TRADES, TRADES with the Asymmetric Negative Contrast (TRADES-ANC), TRADES with the Reverse Attention (TRADES-RA) and TRADES-ANCRA, respectively. As shown in Table~\ref{Tab2}, when incorporating individual ANC or RA, the performance of robustness and generalization has been improved compared with vanilla TRADES. ANC has a larger improvement in clean accuracy than RA, and RA has a better performance on robustness. ANC causes different classes to move away from each other, maintaining sufficiently large margins between categories. This intuitively contributes significantly to generalization. RA scales the feature vectors of natural and adversarial samples based on linear layer weights to align their feature. This enables the model to effortlessly learn a feature distribution that encompasses both natural and adversarial representation, boosting adversarial robustness. Although ANC helps less on robustness than RA, ANC has a larger increase in clean accuracy than that of RA. They complement each other in two aspects. Besides, when TRADES-ANCRA is compared with other methods, the clean accuracy and robust accuracies against all the attacks except FGSM have been enhanced, which indicates that the two strategies are compatible and the combination can alleviate the side effects of independent methods.

\begin{table}[!htbp]
\centering
\caption{Clean and robust accuracy (\%) of ResNet-18 trained by TRADES, TRADES-ANC, TRADES-RA and TRADES-ANCRA against various attacks.}
\label{Tab2}
\begin{tabular}{ccccc}
\toprule
\multicolumn{1}{c}{Defense}&Nat&PGD&FGSM&C\&W\\

\hline
TRADES&78.92&48.40&59.60&47.59\\

TRADES-ANC&80.77&54.18&63.44&49.84\\

TRADES-RA&80.46&61.59&61.48&72.15\\

TRADES-ANCRA&81.70&61.68&61.56&72.36\\

\bottomrule
\end{tabular}
\end{table}

\textbf{hyperparameters} We have used three hyperparameters in the loss function: $\alpha$, $\zeta$ and $\gamma$ . $\alpha$ denotes the weighting factor to adjust the magnitude of the two repulsive forces, which we mentioned in Equation~\ref{eq2} and Equation~\ref{eq3}. $\zeta$ denotes the weight of the asymmetric negative contrast in the total loss, which we mentioned in Equation~\ref{eq4}. We tune these hyperparameters on CIFAR-10 on ResNet-18. $\gamma$ is the weight of auxiliary loss, we set 2.0 as~\cite{CAS}.

\begin{figure}[!htbp]
\vspace{-0.5em}
    \centering
    \includegraphics[width=0.5\textwidth]{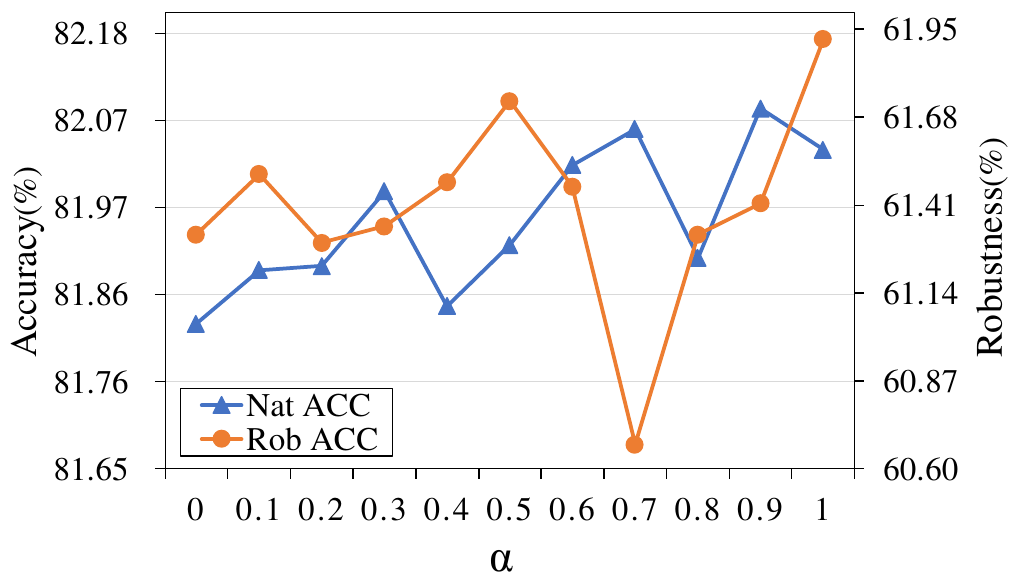}
    \caption{Clean and robust accuracy with different $\alpha$.}
    \label{fig:alpha}
\vspace{-0.5em}
\end{figure}

\begin{figure}[!htbp]
\vspace{-1.0em}
    \centering
    \includegraphics[width=0.5\textwidth]{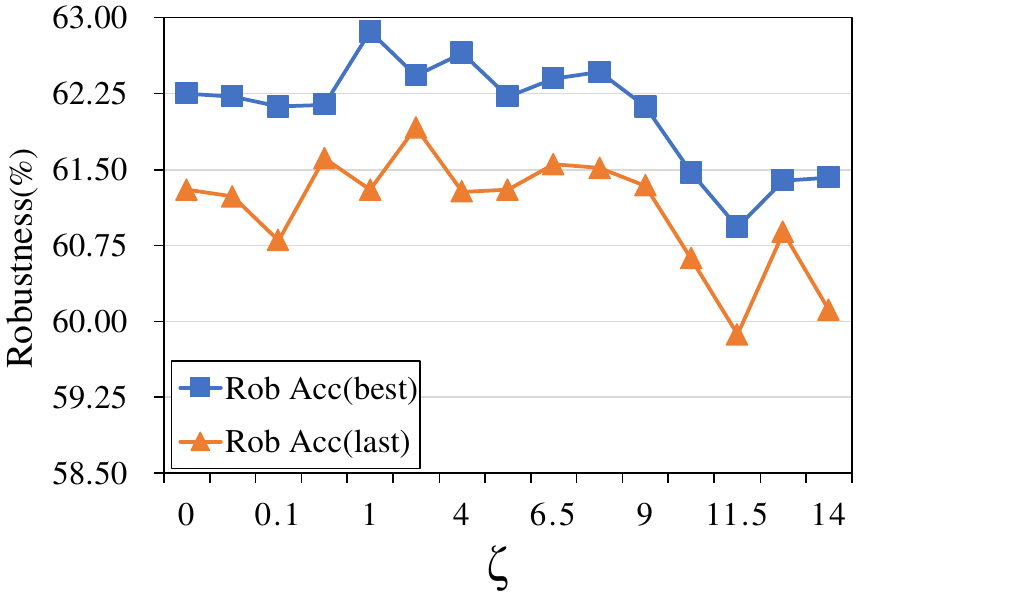}
    \caption{Clean and robust accuracy with different $\zeta$.}
    \label{fig:zeta}
\end{figure}

As shown in Figure~\ref{fig:alpha}, there is a positive relationship between the accuracy and $\alpha$. Though there is an obvious trade-off between the clean and robust accuracy when $\alpha$ equals from 0.5 to 0.7, we can still see an abnormal increasing trend. It is because the larger $\alpha$ leads to the larger repulsive force from the OE to the natural example, to prevent the natural example from being pushed into the wrong class. Besides, as shown in Figure~\ref{fig:zeta}, the robustness has peaked when $\zeta$ equals from 1.0 to 4.0. We choose $\zeta=3.00$ in which models gain the best robust accuracy against PGD-40 at the last epoch.

\textbf{Negative examples strategies}
We compare our strategy (Targeted) with other strategies of negative samples~\cite{ASCL}, including Random, Soft-LS and Hard-LS. To make a comprehensive comparison, we show results of both the best models and the last models with different strategies. As shown in Table~\ref{tb:neg strategies}, our strategy has the best performance of robustness and clean accuracy in the last models, and achieves the best robust accuracy in the best models. Considering that the improvement of our method compared with others is marginal, we also report the training time of different strategies for TRADES-ANCRA to demonstrate another advantage. We train a ResNet-18 model by TRADES with a learning rate of 0.1 to report as a baseline. As shown in Table~\ref{tb:neg strategies}, our strategy costs less time than the average of these selection strategies (9.8 hours) but achieves the best performance. Considering the significant gain in clean and robust accuracy resulting from the proposed method, the cost is relatively worthwhile.

\begin{table}[!htbp]
\setlength\tabcolsep{5.0pt}
\centering
\caption{Results (\%, hour) of four negative examples strategies. Best- and Last- denote the results of the best and last model, respectively. We show the best results with \textbf{bold}.}
\begin{tabular}{cccccc}
\toprule

\multicolumn{1}{c}{Strategy}&Best-Nat&Best-PGD&Last-Nat&Last-PGD&Time\\
\hline

TRADES&82.46&52.17&82.72&51.38&\pmb{6.2}\\

Random&81.44&62.64&81.78&61.71&6.9\\

Soft-LS&82.10&61.83&80.62&58.47&11.4\\

Hard-LS&\pmb{82.30}&62.53&82.13&60.98&11.3\\
Targeted attack&81.36&\pmb{63.08}&\pmb{82.18}&\pmb{62.02}&9.3\\
\bottomrule
\end{tabular}
\label{tb:neg strategies}
\end{table}

\subsection{Limitation} 
Because the weights for reverse attention are determined by predicted classes, the wrong predicted classes may lead to the wrong weighted feature and degraded performance. As shown in Table~\ref{Tab1}, the final predicted results and intermediate predicted labels remain highly consistent. Fortunately, Table~\ref{tab:adaptive} and Figure~\ref{fig:deploy} have indicated that the high dependence on predicted classes does not significantly affect performance. We will further study this limitation and improve it in the future.

 \begin{table}[!htbp]
\vspace{-0.5em}
\centering
\caption{Clean and robust accuracy (\%) of all the probability vectors trained by TRADES-ANCRA. "Final PV wo RA" means we remove reverse attention and then load trained parameters to test it.}
\label{Tab1}
\begin{tabular}{cccc}
\toprule
\multicolumn{1}{c}{Probability Vector (PV)}&Nat&PGD&Adaptive PGD\\
\hline
Auxiliary PV $P$&81.81&83.52&62.25\\

Final PV $P'$&81.81&83.47&62.24\\

Final PV wo RA $P''$&59.77&58.53&52.81\\
\bottomrule
\end{tabular}

\vspace{-0.5em}
\end{table}

 \section{Conclusion} This work addresses the overlook of robust representation learning in adversarial training by a generic AT framework with Exclusion and Alignment criteria. We follow two criteria and propose the asymmetric negative contrast and reverse attention. Specifically, the asymmetric negative contrast based on probabilities freezes natural examples, and only pushes away examples of other classes in the feature space. Besides, the reverse attention weights feature by the parameters of the linear classifier, to provide class information and align feature of the same class. In addition, our framework can be combined with other defense methods in a plug-and-play manner. and be used in the deployment phase without training. It benefits other fields such as representation learning, face recognition and Interpretability of artificial intelligence.


\ifCLASSOPTIONcaptionsoff
  \newpage
\fi

{\small
\bibliographystyle{IEEEtran}
\bibliography{ref}
}

\vfill

\end{document}